\RequirePackage{fix-cm}
\documentclass[smallextended]{svjour3}       
\smartqed  
\usepackage{graphicx}
\usepackage{amsmath}
\usepackage{multirow}
\usepackage{enumerate}
\usepackage{enumitem}
\usepackage{authblk}
\usepackage[sort&compress,numbers]{natbib}
\usepackage[misc]{ifsym}
\usepackage{hyperref}
\usepackage{booktabs}
\usepackage{xcolor}
\usepackage{ulem}
\usepackage{etoolbox}
\usepackage[export]{adjustbox}
\usepackage[symbol*]{footmisc}
\newcommand*{\affaddr}[1]{#1} 
\newcommand*{\affmark}[1][*]{\textsuperscript{#1}}

\begin{document}

\title{An anomaly prediction framework for financial IT systems using hybrid machine learning methods
}

\titlerunning{An anomaly prediction framework for financial IT systems ...}        

\author{\mbox{Jingwen Wang\affmark[1]\footnote[1] \and Jingxin Liu\affmark[1]\footnote[1] \and Juntao Pu\affmark[1]\and
        Qinghong Yang\affmark[1]\and}
        Zhongchen Miao\affmark[2]\and
        Jian Gao\affmark[2]\and
        You Song\affmark[1]$^{\dagger}$}
\authorrunning{Jingwen Wang\and
        Jingxin Liu\and
        Juntao Pu\and
        Qinghong Yang\and
        Zhongchen Miao\and
        Jian Gao\and
        You Song
        }

\institute{
Jingwen Wang \\
\email{wangjingwen@buaa.edu.cn}\\
Jingxin Liu \\
\email{jhljx@buaa.edu.cn} \\
Juntao Pu\\
\email{sy1721108@buaa.edu.cn} \\
Qinghong Yang \\
\email{yangqh@buaa.edu.cn} \\
Zhongchen Miao   \\
\email{miaozc@cffex.com.cn} \\
Jian Gao  \\
\email{gaojian@cffex.com.cn} \\
You Song \\
\email{songyou@buaa.edu.cn}\\
\\
\affaddr{\footnote[1] \,\,Both authors contributed equally to this research.} \\
\affaddr{$^{\dagger}$ You Song is corresponding author.}\\
\affaddr{\affmark[1] School of Software, Beihang University, Beijing, China}\\
\affaddr{\affmark[2] Shanghai Financial Futures Information Technology Co., Ltd}\\
}

\date{Received: date / Accepted: date}

\maketitle
\begin{abstract}
In financial field, a robust IT system is of vital importance to ensure the smooth operation of financial transactions. However, many financial corporations still depend on operators to identify and eliminate the system failures when financial IT systems break down. This traditional operation method is time consuming and extremely inefficient. To improve the efficiency and accuracy of system failure detection and thereby reduce the impact of system failures on financial services, we propose a novel machine learning-based framework to predict the occurrence of system exceptions and failures in a financial IT system. In particular, we first extract rich information from system logs and eliminate noises in the data. Then the cleaned data is leveraged as the input of our proposed anomaly prediction framework which consists of three modules: key performance indicator(KPI) data prediction module, anomaly identification module and severity classification module. Notably, we design a hierarchical architecture of alarm classifiers and try to alleviate the influence of class-imbalance problem on the overall performance. Empirically, the experimental results demonstrate the superior performance of our proposed method on a real-world financial IT system log data set.


\end{abstract}

\section{Introduction}
\label{intro}
IT systems have been widely utilized in a number of areas and have played quite important roles in different corporations. The IT systems in financial companies are busy dealing with large amount of financial transactions every day, providing reliable and effective services to traders. However, system exceptions and failures have tremendously severe and even deadly impact on the financial IT systems, which can affect the smooth operation of financial transactions. In traditional operation methods, the operators need to monitor servers all the time and they suffer a lot from fault location and troubleshooting.  Hence, it is well worth to automatically detect system anomalies and report alarms to operators, which can save time and thereby improve the efficiency of operation.


Recently, extensive research has been conducted on automatically detecting and predicting anomalies in IT systems. \citet{pellegrini2015machine} proposed a machine learning based framework to calculate the remaining time to system failure. However, several procedures require manual intervention and parameters need to be set in advance, which still can not achieve the automatic operation goal. \citet{naveiro2018large} presented a framework aiming to monitor KPI time series and detecting anomalies in a completely automatical way. But the main drawback of this work is the thresholds need be preset empirically.

Although aforementioned methods have been proposed, predicting anomalies in IT systems still faces the following challenges:

\begin{itemize}[label=$\bullet$]
\item \textbf{Inaccuracy of labelling anomalies}: Most of the anomalies are labelled by operators. Different operators may produce different label results for the same sample.
\item \textbf{Incomplete anomaly cases}: System anomalies usually occur rarely, and the collected data set can not cover all kinds of anomalous cases.
\item \textbf{Feature extraction and selection}: The performance of the anomaly prediction model depends on the input features to a large extent, and the raw log files contain lots of redundant information. Hence, it is essential to utilize appropriate features which can accurately capture the characteristics of system anomalies when training the model.
\item \textbf{Class imbalance}: The class imbalance problem means the number of anomalous cases is far exceeded by the number of normal cases in the collected data set. This problem can severely affect the performance of anomaly prediction methods.
\end{itemize}

The first two challenges are inevitable in the data collection process, therefore, we mainly focus on addressing the latter two challenges. As for the feature selection challenge, we combine multiple KPI features from the dirty system logs to depict the system running state as comprehensively as possible. Then we utilize these features to predict the existence of system anomalies. However, in previous works \cite{bertero2017experience,stearley2008bad}, they only focused on text analysis approaches to reveal anomalies, which is prone to be affected by redundant information and noises in the text of system logs. Instead, we do not pay much attention to exploit textual features from system logs. We leverage more expressive KPI features to build the proposed framework.

For the class imbalance problem, it has aroused great attention in both academic and industry. Recently, many imbalanced learning algorithms have been proposed, including sampling methods, cost-sensitive learning methods, kernel-based methods and active learning methods \cite{he2009learning}. In previous work \cite{liu2015opprentice}, it utilized dynamic adjusted threshold method to solve the class imbalance problem, which is sensitive to changes in the data and can affect the performance. In order to reduce the impact of class imbalance problem, we elaborately design a robust multi-level framework by hybrid machine learning methods, in which the KPI data prediction module predicts the KPI features of future time steps, the anomaly identification module distinguishes anomalous samples from normal samples, and the alarm severity classification module finally identifies the severity level for those anomalies. Specifically, in the alarm severity classification module, we also propose a weighted sampling strategy to alleviate the class-imbalance problem in anomalous samples. Hence, the proposed framework is able to reduce the impact of class imbalance problem on the overall performance.

Empirically, we conduct experiments on a real-world system log data set from a financial IT system. The experimental results demonstrate it is effective to utilize KPI time series as key features of the system running state and thereby predict the existence of system anomalies. Furthermore, we analyze the experimental results of different choices of algorithms in each module, further demonstrating the effectiveness and rationality of the propose framework. Note that our proposed framework is an automatic anomaly prediction framework for financial IT systems. When using our framework, the operators do not need to preset parameters, such as thresholds in the existing methods \cite{pellegrini2015machine,naveiro2018large}.

In summary, the main contributions of our paper are as follows.

\begin{itemize}[label=$\bullet$]
\item We propose an anomaly prediction framework which can accurately predict the KPI features and then utilize the predicted features to further predict system anomalies in a totally automatic way.
\item We combine time series data of KPIs and text data of system exception logs to build our framework, instead of directly analyzing the text of system logs.
\item We conduct experiments on a real-world system log data set from a financial IT system, and demonstrate the effectiveness and robustness of our proposed framework.
\item We deploy the framework into the real financial IT system, and prove that our framework can meet the requirements of the real-time anomaly prediction.
\end{itemize}

The rest of the paper is organized as follows. In Sect. \ref{background}, we introduce some preliminary knowledge relevant to the actual server operation business.  In Sect.~\ref{sec:realtedwork}, we review the related work. In Sect.~\ref{sec:method}, we provide a general view of the proposed framework and introduce each module in detail. We report the experimental results in Sect.~\ref{sec:result} and summarize the paper in Sect.~\ref{sec:conclu}.

\section{Background and Preliminaries}
\label{background}
In this section, we first present some essential notions related to system anomaly prediction. Then we introduce the generation process of system log data in the financial IT system.

\subsection{Notions}
In order to detect and predict system exceptions and failures in real applications, system performance indexes are useful measures to reflect the system running state of servers. We define these system performance indexes, like the usage of CPU, disk and memory, as \textbf{key performance indicators}(\textbf{KPIs}). KPI data is then collected and utilized to predict system anomalies. Note that \textbf{system anomalies} refer to system exceptions and failures, rather than anomalies in the time series of KPI data. If a system exception occurs, it would trigger an alarm to remind operators to to resolve it. Hence, it is clear that an \textbf{alarm} is a signal indicating the existence and severity of system anomaly. In other words, predicting system anomalies is equivalent to predicting alarms.

\begin{figure*}[htbp]
\begin{center}
\includegraphics[width=0.6\paperwidth]{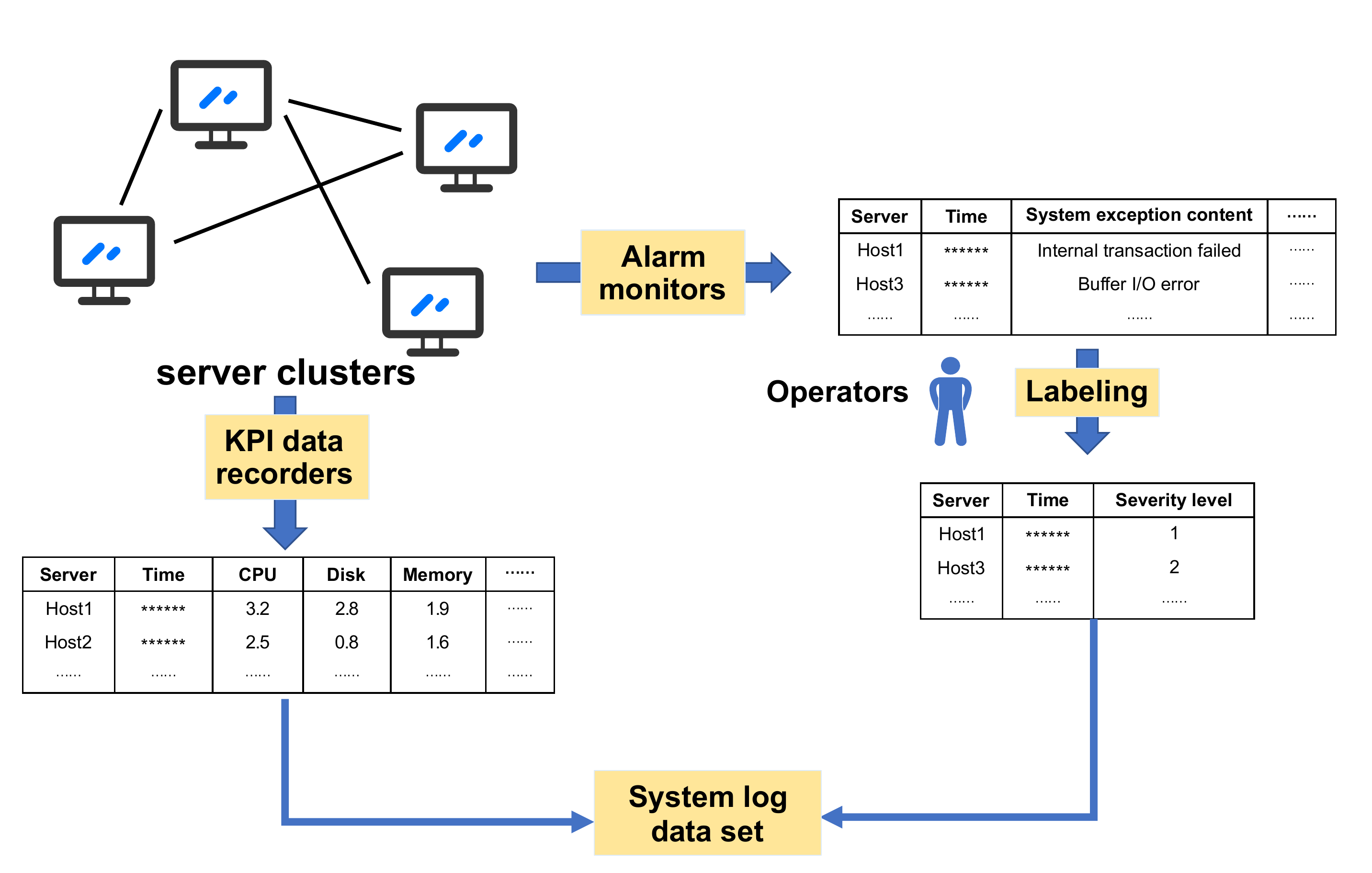}
\caption{The basic background of server operations and the generation process of the system log data set.}
\label{dataset}
\end{center}
\end{figure*}

\subsection{Generation of system logs}

As shown in Fig. \ref{dataset}, server clusters are commonly utilized in financial corporations. While servers are running, KPI data recorders will record the usage of CPU, memory and disks for each server all the time, thereby yielding time series data of KPIs. Meanwhile, if system exceptions or failures occur in some servers, alarm monitors will record detailed information of system anomalies and produce alarms to operators. Then the operators will label the severity level of system anomalies according to the content of corresponding alarms. Therefore, the time series data of server KPIs and the labelled alarm data compose the whole system log data set.

\section{Related work}
\label{sec:realtedwork}
Extensive research has been conducted on detecting and predicting anomalies in IT systems. According to different application scenarios, the type of anomaly may differ from each other. To the best of our knowledge, the anomalies in the previous works can be categorized into two kinds: one is the anomalies in KPI time-series, the other is the anomalies in system log content.

Most of the previous works focus on detecting anomalies in time series. Among all the traditional time-series anomaly detection methods, threshold-based methods are classical and practical ones. \citet{lee2012threshold} proposed a scalable threshold-based solution, called Threshold Compression, to eliminate anomalies and thereby accurately capture the spatial-temporal network dynamics. \citet{liu2015opprentice} utilized dynamic adjusted threshold method to reveal anomalies in time-series data, and deployed this anomaly detection algorithm into a network monitoring system of an Internet-based service.  Some machine learning based methods\cite{salfner2007using,lee2018greenhouse} are also be employed to detect anomalies in time series. \citet{lee2018greenhouse} developed a novel time-series anomaly detection system which combined state-of-the-art machine learning and data management approaches. In addition, other signal processing methods and hybrid methods are demonstrated to be useful in practical applications\cite{lu2009network,laptev2015generic,taylor2018forecasting}. For example, \citet{lu2009network} proposed a new network signal modelling technique based on wavelet analysis technique for detecting anomalies on networks. \citet{laptev2015generic} presented a generic and scalable framework, called EGADS, for automated anomaly detection on large scale time-series data. It utilized blended approaches, including time-series decomposition, change point detection and time-series clustering, to reveal different subtypes of anomaly.

Some other works detect anomalies based on the content of system log data. Text processing and analysis methods are commonly utilized to first extract features in system logs\cite{stearley2008bad}, which ensures the subsequent anomaly detection methods. After the feature extraction process, machine learning methods can be leveraged to distinguish system anomalies. \citet{fulp2008predicting} presented a support vector machine(SVM) based method which distinguished system failures and normal messages based on the frequency of message sequences. \citet{juvonen2014efficient} employed random projection techniques on the preprocessed numeric matrix of system logs and then used Mahalanobis distance to find outliers. Note that unsupervised learning methods are also widely utilized to reveal anomalies \cite{stearley2004towards,hu2018anomaly,du2015behavioral}. For example, \citet{du2015behavioral} presented a new detection method which generated pattern sets based on the effective hierarchical clustering algorithm and detected anomalies according to the relation between the log sequences and the patterns in pattern sets. \citet{hu2018anomaly} designed a similarity clustering algorithm on the system logs and then determined the anomalies according to distance measure.

As for anomaly prediction in IT systems, most of the system anomaly prediction framework consist of several main parts, such as feature selection phase, and prediction model training phase, sometimes including optimal model selection phase. For instance,  \citet{pellegrini2015machine} proposed a machine learning based framework to calculate the remaining time to system failure. It utilized regularization algorithm to select different sets of features, and then predicted system failures based on different generated prediction models. \citet{naveiro2018large} presented a framework with a class of models, and aimed to monitor KPI time series and detect anomalies in a completely automatic way. \citet{sipos2014log} proposed a data-driven approach based on multiple-instance learning for predicting equipment failures by mining equipment event logs. It extracted features from system logs and utilized a bootstrapped feature selection algorithm to select relevant features. After feature selection, it trained a sparse linear classification model by multiple-instance learning and generated predictive scores for test instances.

\section{Methods}
\label{sec:method}

\begin{figure*}[htbp]
\begin{center}
\includegraphics[width=0.55\paperwidth]{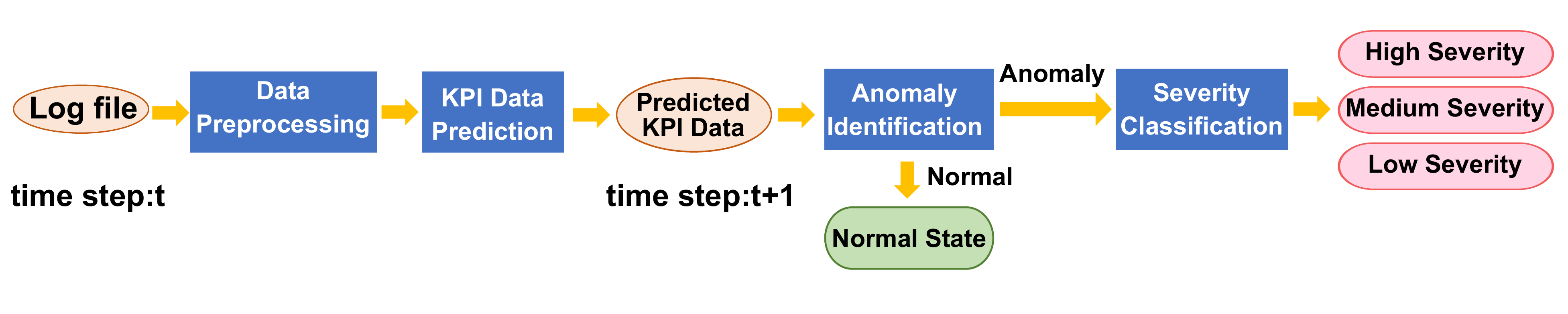}
\caption{The architecture of the Real-Time Anomaly Prediction(RTAP) framework.}
\label{architecture}
\end{center}
\end{figure*}

In this section, we propose a \textit{R}eal-\textit{T}ime \textit{A}nomaly \textit{P}rediction framework, namely RTAP, to predict system anomalies and report severity levels by employing hybrid machine learning methods. The architecture of RTAP includes four modules, which are data preprocessing module, time series prediction module, anomaly identification module and severity classification module, as illustrated in Fig.~\ref{architecture}.

First, the raw system log data at time step $t$ comes as the input of data preprocessing module, which removes data noises and yields cleaned KPI data. Then the KPI data prediction module utilized cleaned KPI data to produce the predicted KPI data at time step $t + 1$. In essence, this module is the core part in our framework, as it determines the accuracy of subsequent severity level prediction results. Once the predicted KPI data at time step $t + 1$ is generated, it is necessary to infer the existence of anomaly and identify the severity level from the predicted KPI data. Due to the data imbalance problem, we design a hierarchical architecture of classifiers, instead of a single multi-class classifier. Note that the anomaly identification module trains two-layer stacking classifiers to infer the existence of anomaly at time step $t + 1$. For those which are discriminated as anomalies, we utilize the severity classification module to identify the severity levels, including low, medium and high severity levels.

\subsection{Data Preprocessing Module}
The raw system log data contains noises and missing KPI values, which can influence the anomaly prediction performance of our proposed framework. In this module, we utilize data standardization, data cleaning and missing value filling techniques to eliminate errors and reduce noise to the most extent.

\subsection{KPI Data Prediction Module}
KPI data prediction module is an essential part of the proposed framework, since the subsequent two modules depend on its output to infer the existence of system anomalies and predict their severity levels. The accuracy of the predicted KPI data at time step $t + 1$ is crucial for the overall performance of the RTAP framework.

Considering the tradeoff between high accuracy and superior real-time capacity, we adopt random forests regression(RFR) \cite{segal2004machine}, a regression version of random forest(RF) \cite{breiman2001random}, as the KPI data predictor. It has shown significant gains in various time series prediction tasks \cite{zarei2013road}. And evaluation results in Sect.~\ref{sec:result} also demonstrate that RFR outperforms other algorithms in this prediction task.

For clarity, we introduce some basic ideas of RF. RF is an ensemble classifier including a number of decision trees. The key idea of RF is to utilize decision trees with low correlation to produce accurate ensemble predictions. Note that there are two important points in the RF: one is that RF leverages a technique, called bagging, to select random samples with replacement in the training set; the other is that when splitting nodes in decision trees, a random subset of features is considered to produce the most appropriate separation, instead of evaluating all possible features at a time. These properties guarantee RF to generalize well to new data and reduce the risk of overfitting. In contrast to RF, RFR utilizes a slightly different cost function of splitting nodes in decision trees. The commonly used node splitting cost function in RFR is the mean square error(MSE) function, instead of information gain or gini coefficient \cite{hssina2014comparative} in RF.

\subsection{Anomaly Identification Module}
As system anomalies usually appear rarely in the server running process and the number of anomalies with high severity level are even fewer, it is challenging to distinguish anomalous cases from normal cases in such class-imbalanced data. Hence, we elaborately design a pipeline to first identify the existence of anomalies and then recognize severity levels for those anomalies, instead of making a multi-classification to discriminate normal cases and all kinds of anomalous cases. The anomaly identification module is the first part of the pipeline, which can also be regarded as a binary classification module.

\begin{figure}[htbp]
\begin{center}
\includegraphics[width=0.5\paperwidth]{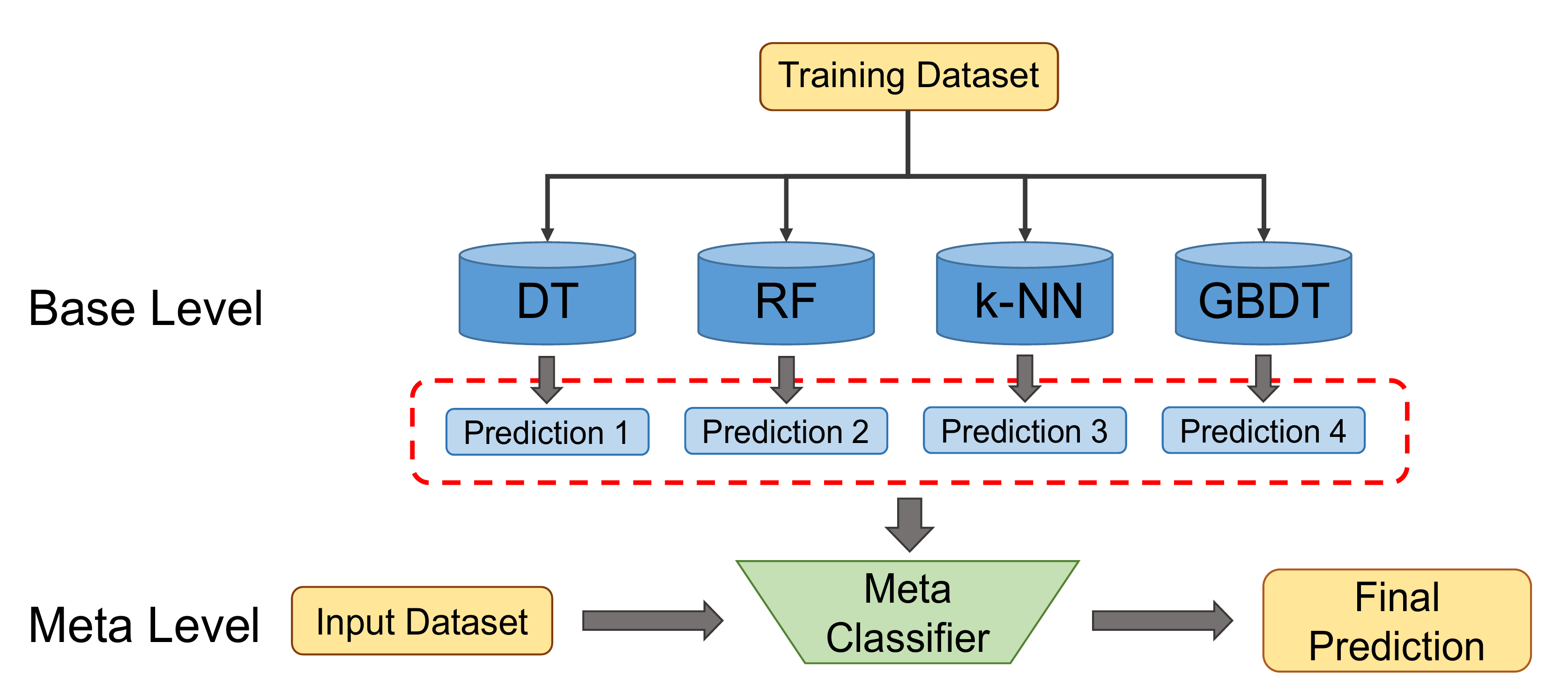}
\caption{The hierarchical structure of stacking classifiers in the anomaly identification module. This structure has two layers: the base layer includes four single classifiers(DT, RF, kNN and GBDT) while the meta layer includes an LR classifier. }
\label{stacking}
\end{center}
\end{figure}


To increase the accuracy of this module, we propose a hierarchical structure involving several classifiers, as illustrated in Fig. \ref{stacking}. The structure contains two layers of classifiers, in which the base layer is comprised of decision tree(DT), random forest(RF), k-Nearest Neighbours(kNN), gradient boosting decision tree(GBDT) \cite{friedman2001greedy,friedman2002stochastic}, and the meta layer is a logistic regression(LR) model blending the output of base classifiers as its input features \cite{sulzmann2011rule}. As follow the literature, simple linear classifier usually works well in the meta layer \cite{witten2016data}. Note that this structure is brought from the stacking method, which is an ensemble technique widely utilized in various applications to improve the performance of classification algorithms \cite{wolpert1992stacked}.

\subsection{Severity Classification Module}
In this module, we only consider how to distinguish system anomalies with different severity levels. We present a k-Nearest Neighbors (kNN)-based method, which is commonly employed in many tasks due to its simplicity and the tolerance in high-dimensional and incomplete data \cite{ban2013referential}.

Formally, in kNN model, given a new KPI sample $x$, the output severity level depends on the k-nearest neighbor samples of $x$ in the training data set. The distance between $x$ and all training samples is based on Euclidean distance, which is defined as:
\begin{equation}
\label{em}
d(x,\hat{x})=\sqrt{\sum_{i=1}^n(x_{i}-\hat{x_{i}})^2}
\end{equation}
where $x$ and $\hat{x}$ represent the two samples and $n$ is the dimension of feature vector $x$.
Let us denote a set $N_{k}(x)$ contains the indices of k-nearest neighbors of the sample $x$, the target value of the alarm level is given by Eq. \ref{knn}. As the severity of system anomalies only has three levels, $k$ is set to 3.
\begin{equation}
\label{knn}
f_{k-NN}(x)=\frac{1}{k}\sum_{i\in N_{k}(x)}y_{i}
\end{equation}

Since anomalies with high severity level are also rare in the system log data, the class-imbalanced problem still needs to be resolved in this module. To alleviate this problem, we propose a weighted sampling strategy before training the kNN model. Anomaly samples are replicated in the training data set according to the weights of different severity levels. In our experiments, we give higher weight to anomalies with higher severity level.

\section{Experiments}
\label{sec:result}

In this section, we first describe some details of the utilized system log data set, and then illustrate the evaluation experiments of each modules and the overall framework. Due to the challenge in obtaining source codes of similar anomaly prediction frameworks, we do not make any comparison between the RTAP framework and others. However, we provide insights on how each module is indispensable in the RTAP framework by performing detailed comparison on various design choices of our framework.

\subsection{Data Set}
The data set utilized in this paper is a part of the actual system log files of server clusters in a financial company. And the system logs come from four different types of financial business servers. For ease of presentation, we denote these types as \texttt{Biz}, \texttt{Mon}, \texttt{Ora} and \texttt{Trd}, respectively. More specifically, the system logs contain two types of data: one is KPI time series which record system performance states of 258 servers for approximate 5 months, the other is alarm log data which record the alarm status of the entire system in detail, including server name, alarm time, alarm content, alarm severity level, etc. After data preprocessing stage, the cleaned data contains approximate 640000 records of KPI time series in total, but only contains about 9000 alarm records.


To exploit abundant information from system logs and thereby make better anomaly prediction, we elaborately select several features, such as the maximum, minimum and average of CPU usage, memory usage, and usage of several disks, to depict the running state of servers at a certain time. Due to the technology limitations at the period of collecting system logs, the obtained system logs are coarse-grained time-series data in which the minimal time interval is one hour. Hence, we also include the KPI features of previous hours in the training process.

\subsection{Evaluation Metrics}
In our experiments, we evaluate the performance of each module in the RTAP framework and the overall performance of the RTAP framework. For KPI data prediction module performance evaluation, we use root mean squared error (RMSE) as the evaluation metrics to measure the time-series prediction error. For clarity, we give the definition of RMSE as shown in Eq. \ref{e3}:

\begin{equation}
{\rm RMSE} = \sqrt{\frac{1}{n}\sum_{i=1}^n(r_{i}-\hat{r_{i}})^2}
\label{e3}
\end{equation}
where $n$ is the sample size, $r_{i}$ and $\hat{r_{i}}$ are the predicted KPI value and the real KPI value.

For experiments in the anomaly identification module, we evaluate the binary-class classification results by three evaluation metrics: precision, recall and $F_{\beta}$ score, which are given by:

\begin{equation}
\begin{aligned}
\mathrm{F_{\beta}\;score} &= \frac{(1+\beta^2) \times \mathrm{precision} \times \mathrm{recall}}{(\beta^2 \times \mathrm{precison}) + \mathrm{recall}} \\
\mathrm{precision} &= \frac{\mathrm{TP}}{\mathrm{TP} + \mathrm{FP}} \\
\mathrm{recall} &= \frac{\mathrm{TP}}{\mathrm{TP} + \mathrm{TN}}
\end{aligned}
\end{equation}
where $\mathrm{TP}, \mathrm{FP}$ and $\mathrm{TN}$ refer to true positive, false positive and true negative sample numbers. We choose $F_{0.5}$ score as the final evaluation metrics, as the precision often outweighs recall under the actual server operation and maintenance scenario.

For other experiments, we use Macro $F_{\beta}$ score and Micro $F_{\beta}$ score as the evaluation metrics to evaluate the multi-class classification results. Macro $F_{\beta}$ score is a metric which gives equal weight to each class. Its definition is as follows:
\begin{equation}
{\rm Macro}\;F_{\beta}\;{\rm score} = \frac{\sum_{A \in \mathcal{C}}F_{\beta}(A)}{|\mathcal{C}|}
\end{equation}
where $\mathcal{C}$ is the overall label set and $F_{\beta}(A)$ is the $F_{\beta}$ score of label $A$. Micro $F_{\beta}$ score is a metric which gives equal weight to each instance. Its definition is as follows:
\begin{equation}
\begin{aligned}
{\rm Micro}\;F_{\beta}\;{\rm score} &= \frac{(1+\beta^2) \times \mathrm{precision'} \times \mathrm{recall'}}{(\beta^2 \times \mathrm{precison'}) + \mathrm{recall'}} \\
\mathrm{precision'} &= \frac{\sum_{A \in \mathcal{C}} \mathrm{TP(A)}}{\sum_{A \in \mathcal{C}} \mathrm{TP(A)} + \mathrm{FP(A)}} \\
\mathrm{recall'} &= \frac{\sum_{A \in \mathcal{C}} \mathrm{TP(A)}}{\sum_{A \in \mathcal{C}} \mathrm{TP(A)} + \mathrm{TN(A)}}
\end{aligned}
\end{equation}
As aforementioned, we choose Macro $F_{0.5}$ score and Micro $F_{0.5}$ score as the final evaluation metrics.


\subsection{Evaluation of KPI Data Prediction Module}

In this experiment, several traditional methods including support vector regression(SVR), moving average (MA) and exponential smoothing(ES) are utilized for comparison. To objectively evaluate the performance of each method, we select a naive method as baseline: setting the real KPI value at time step $t$ as the predicted KPI value at time step $t + 1$. As the baseline is a quite weak predictor, the performance comparison to it can reflect the actual prediction capacity of each method.

\begin{table}[htbp]
\linespread{1}
\begin{center}
\small
\caption{Comparison of RMSE values among all compared KPI data prediction methods. The predicted features include maximum, minimum and average of KPI features.}
\begin{tabular}{@{}llccc@{}}
\toprule
Business Type & Method & Maximum & Minimum & Average\\ \midrule
Biz & baseline & 1.62 & 0.33 & 0.22 \\
 & MA & 1.35 & 0.37 & 0.35 \\
 & ES & 1.46 & 0.31 & 0.23 \\
 & SVR & 1.29 & 0.31 & 0.24 \\
 & \textbf{RFR} & \textbf{1.31} & \textbf{0.33} & \textbf{0.26} \\
 \\
Mon & baseline & 1.00 & 0.23 & 0.17 \\
 & MA & 0.82 & 0.30 & 0.31 \\
 & ES & 0.88 & 0.22 & 0.17 \\
 & SVR & 1.06 & 0.62 & 0.50 \\
\textbf{} & \textbf{RFR} & \textbf{0.89} & \textbf{0.27} & \textbf{0.20} \\
\\
Ora & baseline & 1.30 & 0.10 & 0.09 \\
 & MA & 1.04 & 0.15 & 0.15 \\
 & ES & 1.17 & 0.09 & 0.08 \\
 & SVR & 1.11 & 0.11 & 0.10 \\
\textbf{} & \textbf{RFR} & \textbf{1.07} & \textbf{0.10} & \textbf{0.11} \\
\\
Trd & baseline & 0.20 & 0.09 & 0.07 \\
 & MA & 0.21 & 0.14 & 0.14 \\
 & ES & 0.18 & 0.09 & 0.08 \\
 & SVR & 0.17 & 0.10 & 0.09 \\
\textbf{} & \textbf{RFR} & \textbf{0.16} & \textbf{0.09} & \textbf{0.09} \\ \bottomrule
\end{tabular}
\label{timeseries}
\end{center}
\end{table}

Then we evaluate the KPI data prediction performance for all compared methods, as illustrated in Table.~\ref{timeseries}. We can observe that RFR has less prediction errors than other methods over three KPI features (maximum, minimum and average). Note that RFR consistently outperforms other methods across four business types, which demonstrates RFR is an effective and robust method to predict KPI data of this data set.

\begin{figure*}[htbp]
\begin{center}
\includegraphics[width=1\textwidth]{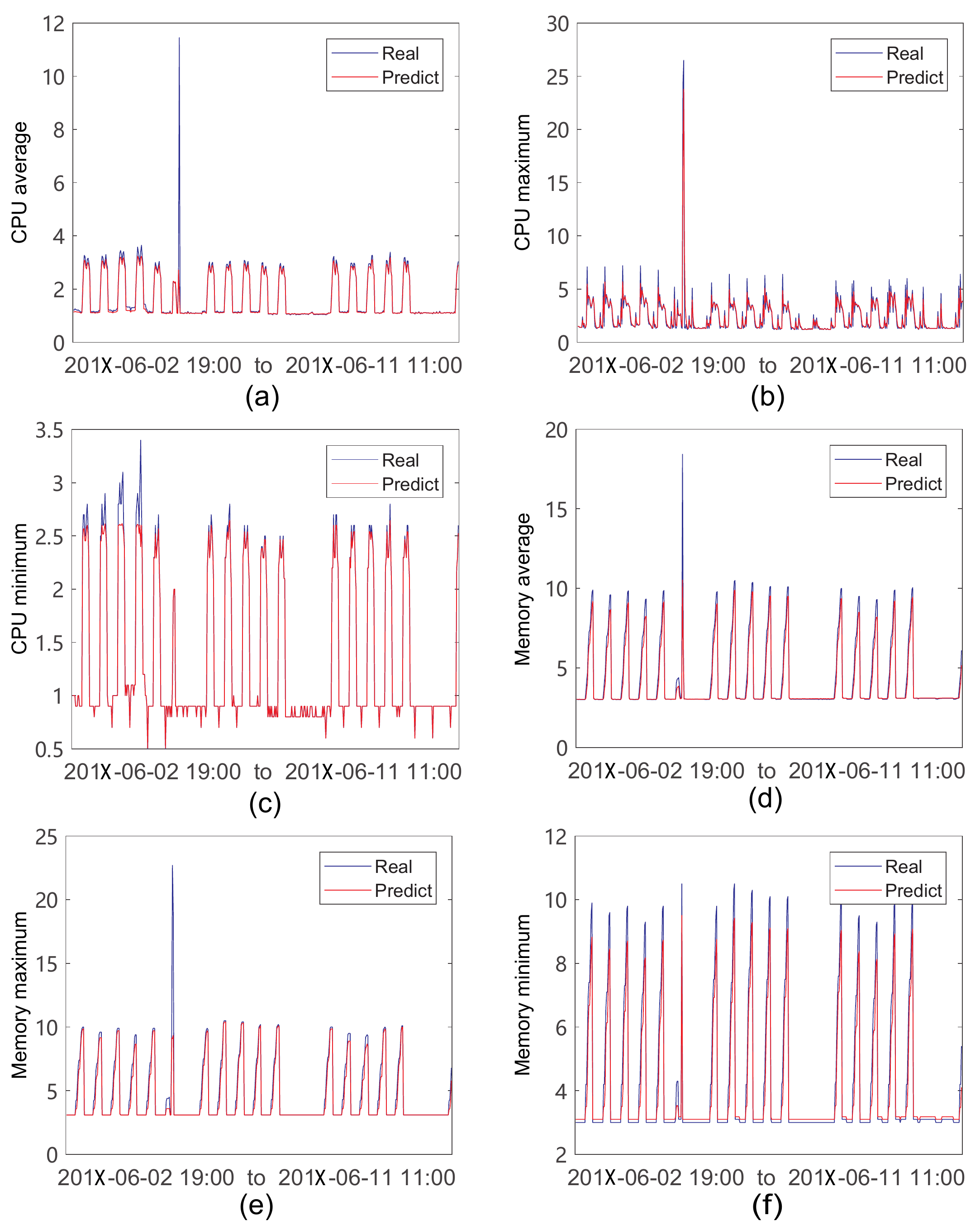}
\caption{The KPI data prediction results of the RFR model. For simplicity, the KPI features shown here only include the maximum and average of CPU and memory usage.}
\label{timepre}
\end{center}
\end{figure*}

To further evaluate the prediction accuracy of RFR over timestamps, we select a server called \texttt{alarmsvr1} as an example to evaluate the long-term time-series prediction performance. Fig.~\ref{timepre} shows the real KPI data curves and the corresponding predicted KPI data curves. It indicates that RFR performs well on the prediction of each feature, since RFR not only effectively captures the trends of KPI data but also predicts the KPI values accurately\footnote{Exact date information cannot be provided due to the protection of business information.}.

\subsection{Evaluation of Anomaly Identification Module}

\begin{table}[htbp]
\linespread{1}
\small
\begin{center}
\caption{The classification results between the stacking classifier we proposed and its base classifiers.}
\begin{tabular}{@{}llccc@{}}
\toprule
Business & Method & Precision & Recall & ${F_{0.5}}\,{\rm Score}$\\ \midrule
Biz & DT & 0.6906 & 0.7319 & 0.6985 \\
 & RF & 0.8744 & 0.7080 & 0.8351 \\
 & k-NN & 0.7401 & 0.6290 & 0.7300 \\
 & GBDT & 0.8303 & 0.6206 & 0.7777 \\
 & \textbf{Stacking} & \textbf{0.8803} & \textbf{0.7017} & \textbf{0.8376} \\
 \\
Mon & DT & 0.5707 & 0.6660 & 0.5875 \\
 & RF & 0.8792 & 0.6427 & 0.8189 \\
 & k-NN & 0.7323 & 0.5373 & 0.6827 \\
 & GBDT & 0.8568 & 0.4339 & 0.7170 \\
\textbf{} & \textbf{Stacking} & \textbf{0.8754} & \textbf{0.6961} & \textbf{0.8325} \\
\\
Ora & DT & 0.4482 & 0.5731 & 0.4686 \\
 & RF & 0.7799 & 0.4901 & 0.6974 \\
 & k-NN & 0.5658 & 0.2549 & 0.4548 \\
 & GBDT & 0.6133 & 0.2194 & 0.4513 \\
\textbf{} & \textbf{Stacking} & \textbf{0.8511} & \textbf{0.5653} & \textbf{0.7729} \\
\\
Trd & DT & 0.4109 & 0.5887 & 0.4373 \\
 & \textbf{RF} & \textbf{0.8621} & \textbf{0.5319} & \textbf{0.7669} \\
 & k-NN & 0.8103 & 0.3333 & 0.6300 \\
 & GBDT & 0.7282 & 0.5319 & 0.6781 \\
\textbf{} & Stacking & 0.8542 & 0.5190 & 0.7564 \\ \bottomrule
\end{tabular}
\label{table2}
\end{center}
\end{table}

All the algorithms are trained and tested on four types of business data which are \texttt{Biz}, \texttt{Mon}, \texttt{Ora} and \texttt{Trd}, respectively.  Table.~\ref{table2} reflects the overall performance statistics of stacking classifier and other base classifiers. We can observe the stacking classifier outperforms other algorithms in most cases, although RF performs best on the \texttt{Trd} data. It indicates that the stacking classifier can distinguish normal cases and anomalous cases more accurately than its base classifiers. In addition, the stacking classifier behaves more stably on various types of business data, with all $F_{0.5}$ scores greater than 0.75. Note that the test data utilized in this module include normal samples and anomalous samples with a proportion of $70: 1$ in \texttt{Biz}, $60: 1$ in \texttt{Mon}, $40: 1$ in \texttt{Ora} and $275:1$ in \texttt{Trd}. The results in Table.~\ref{table2} suggest the stacking classifier can effectively reduce the false positive rate in such highly imbalanced data. Overall, we can conclude that our proposed stacking classifier is robust on all types of business data, and can predict the existence of system anomalies accurately based on the predicted KPI features.

\subsection{Evaluation of Severity Classification Module}

In this module, we evaluate the effectiveness of our proposed weighted sampling strategy. We compare the performance of the kNN model with weighted sampling strategy and its counterpart without the sampling strategy. As illustrated in Table.~\ref{sampling}, we can see that when the sampling strategy is utilized, the kNN model consistently performs better on all four types of business data. We also observe that the kNN model with sampling strategy achieves a gain of at least $0.2$ in Macro $F_{0.5}$ score, which suggests this weighted sampling strategy can alleviate the impact of class-imbalance problem on the performance of this module. It further demonstrates this module can distinguish anomalies with different severity levels in a more accurate way.

\begin{table}[htbp]
\linespread{1}
\small
\begin{center}
\caption{The classification results of whether using the weighted sampling strategy in the kNN model or not. The test data utilized in this table contains system anomalies with all three severity levels.}
\begin{tabular}{@{}lccc@{}}
\toprule
Business & Sampling & ${\rm Macro\;}F_{0.5}\,{\rm Score}$ & ${\rm Micro\;}F_{0.5}\,{\rm Score}$ \\ \midrule
Biz & \textbf{Yes} & \textbf{0.7680} & \textbf{0.9940} \\
 & No & 0.5272 & 0.9924 \\
 \\
Mon & \textbf{Yes} & \textbf{0.7102} & \textbf{0.9941} \\
 & No & 0.4998 & 0.9876 \\
\\
Ora & \textbf{Yes} & \textbf{0.6305} & \textbf{0.9874} \\
 & No & 0.4482 & 0.9750 \\
\\
Trd & \textbf{Yes} & \textbf{0.8591} & \textbf{0.9979} \\
 & No & 0.5607 & 0.9976 \\ \bottomrule
\end{tabular}
\label{sampling}
\end{center}
\end{table}


\subsection{Evaluation of the RTAP Framework}
To evaluate the overall performance of our RTAP framework, we utilize the data from 201X-01-01 00:00 to 201X-05-31 23:00 \footnote{Exact date information cannot be provided due to the protection of business information.} to train the framework and test its performance on the data from 201X-06-01 00:00 to 201X-06-30 23:00. As RF is a commonly used ensemble classifier which is robust to noises and performs well in many scenarios, we replace the anomaly identification module and the subsequent severity classification module with a single RF classifier as a comparable baseline method, namely RTAP-C.

As shown in Table. \ref{table666}, we can see that both RTAP and the baseline can reveal all normal cases in the test data. However, RTAP almost consistently outperforms the baseline on anomalous cases across all types of business data. Specifically, RTAP can predict high level anomalies quite accurately in \texttt{Biz} data, while the baseline can not even predict the existence of high level anomalies in \texttt{Biz} and \texttt{Ora} data. It suggests that our framework can effectively alleviate the class imbalance problem in the system log data and thereby produce accurate results when predict different severity levels of system anomalies.

\begin{table}[htbp]
\linespread{1}
\small
\begin{center}
\caption{The overall performance between the RTAP framework and the corresponding baseline method. The table illustrates the $F_{0.5}$ score on each type of data: normal cases and three severity levels of anomalous cases, as well as the macro $F_{0.5}$ score and the micro $F_{0.5}$ score. The line symbol in the table mean that the test data do not include corresponding types of cases.}
\begin{tabular}{@{}llcccccc@{}}
\toprule
Business & method & normal & low & medium & high & macro & micro\\ \midrule
Biz & RTAP-C & 0.9955 & 0.6559 & 0.4301 & 0.0000 & 0.5118  & 0.9904 \\
 & \textbf{RTAP} & \textbf{0.9963} & \textbf{0.8107} & \textbf{0.2476} & \textbf{0.8333} & \textbf{0.6644}  & \textbf{0.9939} \\
 \\
Mon & RTAP-C & 0.9945 & 0.6208 & 0.3508 & 0.1999 & 0.5091  & 0.9874 \\
 & \textbf{RTAP} & \textbf{0.9961} & \textbf{0.8580} & \textbf{0.3845} & \textbf{0.4545} & \textbf{0.6121} & \textbf{0.9940}  \\
\\
Ora & RTAP-C & 0.9900 & 0.4968  & 0.6250 & 0.0000  & 0.4827 &  0.9759 \\
 & \textbf{RTAP} & \textbf{0.9920} & \textbf{0.7536} & \textbf{0.8333} & \textbf{0.4762} & \textbf{0.6711} & \textbf{0.9876} \\
\\
Trd & RTAP-C & 0.9981 & 0.5960 & ------ & ------ & 0.4917 & 0.9977 \\
 & \textbf{RTAP} & \textbf{0.9984} & \textbf{0.6219} & ------ & ------ & \textbf{0.7227} & \textbf{0.9976} \\ \bottomrule
\end{tabular}
\label{table666}
\end{center}
\end{table}



\section{Conclusion}
\label{sec:conclu}

In this paper, we propose an anomaly prediction framework to predict the existence and the severity level of anomalies in financial IT systems. Compared with previous works, we combine time series data of KPIs and text data of system exception logs to build the anomaly prediction framework. To select useful features from raw data, we integrate a number of KPI features to depict the system running state and further predict system anomalies. As class imbalance problem would tremendously affect the overall performance, we design a hierarchical architecture of alarm classifiers and utilize effective data sampling strategy to alleviate the problem. Empirically, the experimental results demonstrate the effectiveness and rationality of the propose framework. The results also give some insights for extending the framework into other financial IT systems, as long as similar KPI features can be recorded in the data collection process.

Due to the technology limitations at the period of collecting system logs, one of the limitations of this work is the obtained KPI time series are coarse-grained. Therefore, KPI changes in small time intervals can not be captured in our framework. Our future work is to optimize the KPI data prediction model on the fine-grained KPI time series, and make the proposed framework predict anomaly severity levels more accurately.

\section{Acknowledgement}
This work is supported by Shanghai Financial Futures Information Technology Co., Ltd. We acknowledge Shanghai Financial Futures Information Technology for proposing the research demand of predicting server anomalies based on time series data of server KPIs. We also thank the reviewers for their careful reading and insightful comments on our manuscript.



\end{document}